\documentclass[journal]{IEEEtran}
\ifCLASSINFOpdf
\else
\fi
\usepackage{cite}
\usepackage{subfig}
\usepackage{amsmath,amssymb,amsfonts}
\usepackage{dsfont}
\usepackage{algorithmic}
\usepackage{algorithm}
\usepackage{graphicx}
\usepackage{textcomp}
\usepackage{xcolor}
\usepackage{lipsum}
\usepackage{enumitem}
\usepackage{tabularx,booktabs}
\newcolumntype{C}{>{\centering\arraybackslash}X} % centered version of "X" type
\usepackage{subfig}

\begin{document}
\title{A Novel Monte-Carlo Compressed Sensing and Dictionary Learning Method for the Efficient Path Planning of Remote Sensing Robots}

\author{Alghalya Al-Hajri, Ejmen Al-Ubejdij, Aiman Erbad,~\IEEEmembership{Senior Member,~IEEE}, Ali~Safa,~\IEEEmembership{Member,~IEEE}
% <-this % stops a space
\thanks{A. Al-Hajri, E. Al-Ubejdij and A. Safa are with the College of Science and Engineering, Hamad Bin Khalifa University, Doha, Qatar.}% <-this % stops a space
\thanks{A. Erbad is with the College of Engineering, Qatar University, Doha, Qatar.}
\thanks{A. Safa supervised the work as Principal Investigator, proposed the presented methods, contributed to the technical developments, as well as to the writing of the manuscript. A. Al-Hajri and E. Al-Ubejdij contributed to the technical developments and to the writing of the manuscript. A. Erbad contributed to the writing of the manuscript.}% <-this % stops a space
%\thanks{Manuscript received April 19, 2005; revised September 17, 2014.}
}


\markboth{Journal of \LaTeX\ Class Files,~Vol.~13, No.~9, September~2014}%
{Shell \MakeLowercase{\textit{et al.}}: Bare Demo of IEEEtran.cls for Journals}



\maketitle

\begin{abstract}
In recent years, Compressed Sensing (CS) has gained significant interest as a technique for acquiring high-resolution sensory data using fewer measurements than traditional Nyquist sampling requires. At the same time, autonomous robotic platforms such as drones and rovers have become increasingly popular tools for remote sensing and environmental monitoring tasks, including measurements of temperature, humidity, and air quality. Within this context, this paper presents, to the best of our knowledge, the first investigation into how the structure of CS measurement matrices can be exploited to design optimized sampling trajectories for robotic environmental data collection. We propose a novel Monte Carlo optimization framework that generates measurement matrices designed to minimize both the robot’s traversal path length and the signal reconstruction error within the CS framework. Central to our approach is the application of Dictionary Learning (DL) to obtain a data-driven sparsifying transform, which enhances reconstruction accuracy while further reducing the number of samples that the robot needs to collect. We demonstrate the effectiveness of our method through experiments reconstructing $NO_2$ pollution maps over the Gulf region. The results indicate that our approach can reduce robot travel distance to less than $10\%$ of a full-coverage path, while improving reconstruction accuracy by over a factor of five compared to traditional CS methods based on DCT and polynomial dictionaries, as well as by a factor of two compared to previously-proposed Informative Path Planning (IPP) methods.

%The abstract goes here. This paper discusses for the first time how the structure of the measurement matrices found in Compressed Sensing can be exploited in order to derive robot data acquisition paths.
\end{abstract}


% \fi
\IEEEpeerreviewmaketitle

%\section*{Supplementary Material}
%Code is provided as open source at: \textcolor{blue}{put url, CITE POLYNOMIAL}

\section{Introduction}
%\textcolor{blue}{cite polynomial}
\IEEEPARstart{T}{he} use of autonomous robots for remote environmental sensing has gained a high attention in the past decades, from autonomous underwater vehicles (AUVs) for oceanic pollution analysis \cite{ocean1, ocean2} and drones equipped with air quality sensors \cite{air1, air2}, to data collection rovers \cite{rover1} such as NASA's Curiosity robot operating on Mars \cite{rover2}. Using robots instead of static sensors for environmental data collection presents many advantages such as on-demand deployability and an increased sensor placement flexibility. On the other hand, acquiring high-spatial-resolution data using robots such as rovers and drones can be challenging due to their limitations in terms of acquisition speed, battery life and overall operation time. For example, if a drone needs to be used for collecting environmental data across a vast piece of land with high spatial resolution, a significantly long data collection path will need to be followed by the robot as the drone navigates the environment in a grid-like fashion. In addition to the limited drone battery life which limits the grid path length that the robot can follow (and hence, the spatial resolution), following a standard grid-like sampling pattern could make the data collection time unrealistically long in many scenarios, jeopardizing the acquisition of sensory data with high spatial resolution \cite{infopathplan, infogad}. 
\begin{figure}[t]
    \centering
    \includegraphics[width=0.53\textwidth]{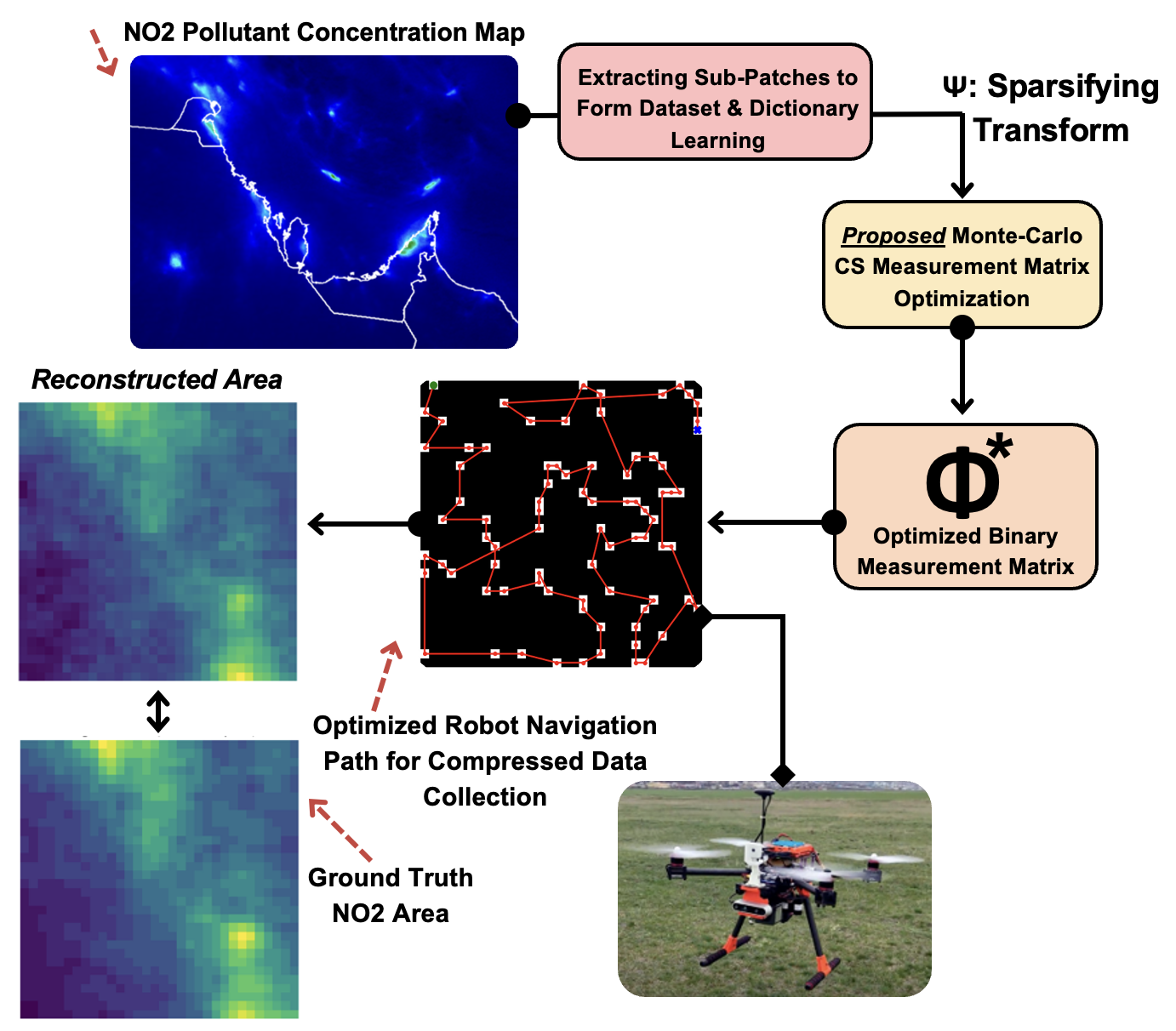} % Adjust the width percentage
    \caption{\textit{\textbf{Proposed robot path planning approach for the compressed sensing (CS) of spatial environmental signals.} }}
    \label{graphabstract}
\end{figure}

To help alleviate these issues and enable high-spatial-resolution data sampling with resource-constrained robots, this paper proposes a novel Monte-Carlo algorithm for robot path planning based on the structure of random measurement matrices $\Phi$ used within a Compressed Sensing (CS) framework \cite{cspaper} (see Fig. \ref{graphabstract}). CS is an innovative sensing paradigm that has attracted much attention in the past decade, which enables the recovery of higher resolution natural signals (e.g., of dimension $N$) using a reduced set of $M\ll N$ measurements \cite{cspaper2}. Hence, the sub-Nyquist sampling ability of CS makes it well suited for our given robot sensing application. Our proposed method generates random binary measurement matrix samples and evaluates their fitness based on the obtained robot measurement path length and the incoherence between the measurement matrix sample $\Phi$ and the sparsity-inducing change of basis matrix $\Psi$ within the CS framework. Crucially, our method does not make use of traditional \textit{pre-defined} sparsity-inducing basis $\Psi$ (such as Fourier and polynomial basis \cite{polycs}) but rather, seeks to \textit{learn} an optimal change of basis $\Psi$ from the data source using a Dictionary Learning (DL) approach \cite{dictionarylearning}. As shown in this work, DL enables the learning of a change of basis $\Psi$ that minimizes the signal recovery error while enabling the use of highly-sparse measurement matrices $\Phi$, leading to shorter robot path lengths compared to the use of pre-defined change of basis. The contributions of this paper are the following:
\begin{enumerate}
    \item We introduce what is, to the best of our knowledge, a novel Monte-Carlo sampling method for the path planning of data collection robots based on the structure of the measurement matrix $\Phi$ within a CS framework.
    \item We experimentally demonstrate the usefulness of the proposed approach by applying our method for the recovery of $NO_2$ pollution maps within the Gulf region. 
    \item Our experimental results demonstrate a significant reduction of the robot data acquisition path length to less than $10\%$ of what is normally required for total area coverage, while leading to high-precision signal recovery, with more than $\times 5$ lower signal recovery errors compared to the use of standard Discrete Cosine Transform (DCT) and polynomial-based CS procedures. 
    \item We compare our proposed approach to previously-proposed Informative Path Planning (IPP) methods \cite{infopathplan, infogad} and we demonstrate that our approach achieves more than $\times 2$ lower map recovery error for a given robot path length, while being significantly less computationally complex since \textit{online} optimization is not required, as opposed to the IPP case.
\end{enumerate}

This paper is organized as follows. Background is provided is Section \ref{back}. Our method is detailed in Section \ref{method}. Experimental results are shown in Section \ref{results}. Conclusions are provided in Section \ref{conclus}.

\section{Background}
\label{back}

\subsection{Prior Work}
\label{priorwork}
Compressed sensing (CS) has been applied in various environmental sensing systems to reduce the number of physical measurements needed to reconstruct spatial fields. While most of these methods focus on improving reconstruction accuracy under sparse sampling, very few consider the implications of how the structure of the measurement matrix can influence or be aligned with robot motion planning, which becomes especially important in real-world deployments involving mobile platforms.
M. T. Nguyen \textit{et al.} \cite{nguyen2020} proposed a distributed compressed sensing method for multi-robot environmental monitoring. Robots moved using a random walk strategy and shared their measurements locally, allowing each robot to perform field reconstruction without relying on a central unit. Although the system reduces communication overhead and supports mobility, the robot paths are not optimized, and the measurement strategy is not learned or adjusted for reconstruction efficiency. S. Yan \textit{et al.} \cite{yan2012} explored static sensor deployment for field reconstruction using a Gaussian kernel basis and Bayesian CS. Their method focused on reducing the number of deployed sensors through innovative sampling. Still, the framework assumes fixed sensing locations and a predefined sparse basis, without considering adaptive measurement matrix design or mobility.
Deep learning has also been integrated into CS-based environmental applications. For instance, Q. Chang \textit{et al.} \cite{chang2023} introduced a GAN-based model to impute missing values in urban environmental datasets, demonstrating improvements in reconstruction under high data loss. However, the data is collected from fixed-grid sensors, and the method assumes no control over sampling locations or sensing trajectories. Similarly, S. Liu \textit{et al.} \cite{liu2024} combined compressed sensing with GANs for visible light-based robot localization. While the paper is relevant in showing the use of CS in robotics, the focus is on cooperative localization, not environmental reconstruction, and no path planning or matrix optimization is involved.

To the best of our knowledge, none of these existing methods study how the structure of the CS measurement matrix can be optimized as a way to influence robot path planning. This is one of the important gaps that our work addresses in this paper.

More closely related to our work, the Informative Path Planning (IPP) method has been proposed by R. Marchant \textit{et al.} \cite{infopathplan} and G. Hollinger \textit{et al.} \cite{infogad} for recovering environmental maps in which spatial correlations can be found. IPP-based methods work as follows. First, the environment sensing robot initializes a Gaussian Process model for the environment, effectively modeling the map to be recovered as a mixture of Gaussian functions. Then, the robot seeks to move in data sampling directions that maximize the \textit{information gain} of the Gaussian Process, defined as the ratio between the prior Gaussian model and the posterior Gaussian model obtained after integrating the new data point. Hence, the movement of the robot is decided online in real time, necessitating a compute-expensive gradient descent step per robot sensing step \cite{infopathplan}. At the end of the online path planning process, a Gaussian Process model is obtained representing the map to be recovered by the robot. 

Even though the goal of our proposed method is the same as IPP (i.e., recovering environmental maps from few sensing data), our proposed algorithm utilizes a fundamentally different approach for robot path planning. First, we make use of Compressed Sensing and Dictionary Learning instead of Gaussian Processes. Second, the robot path optimization is formulated as an offline CS sensing matrix optimization problem in our case, which drops the need for compute-expensive online path planning as in IPP. Since IPP constitutes the main comparable method previously proposed in literature, we compare the performance of our method against IPP in Section \ref{results}.

\subsection{Background on CS and Dictionary Learning}
\label{techback}
\subsubsection{Compressed Sensing (CS)} is concerned with the sensing or \textit{recovery} of an $N$-dimensional signal $\Bar{s}$ from a small number $M \ll N$ of measurements. The original signal that needs to be recovered $\Bar{s}$ is measured through a $M\times N$ measurement matrix $\Phi$ as:
\begin{equation}
    \Bar{s}_m=\Phi \Bar{s}
    \label{meass}
\end{equation}

Furthermore, it is assumed that the original signal $\Bar{s}$ can be represented as a \textit{sparse} set of coefficients $\Bar{c}$ in another basis, through a change of basis $\Psi$, meaning that $\Bar{c}$ is a $N$-dimensional vector with only a few non-zero entries, and is related to $\Bar{s}$ as follows:
\begin{equation}
    \Bar{s} = \Psi \Bar{c}
\end{equation}

Then, it can be shown that the original signal $\Bar{s}$ can be recovered with arbitrary precision from the limited set of $M$ measurements $\Bar{s}_m$ in (\ref{meass}) by solving the following optimization problem:
\begin{equation}
    \Bar{c}^* = \arg \min_{\Bar{c}} \frac{1}{2}||\Phi \Psi \Bar{c} - \Bar{s}_m||_2^2 + \lambda || \Bar{c} ||_1
    \label{cseq}
\end{equation}
where $\lambda$ is a hyperparameter setting the sparsity level of $\Bar{c}$ through the sparsity-inducing $l_1$-norm penalty $|| \Bar{c} ||_1$. After (\ref{cseq}) has been solved using e.g., the iterative soft-thresholding algorithm (ISTA), the original signal can be recovered with arbitrary precision as: 
\begin{equation}
    \Bar{s} \approx \Psi \Bar{c}^*
\end{equation}

It is important to note that the choice of the basis matrix $\Psi$ has a strong impact on the tradeoff between the number of measurements $M$ and the signal recovery quality, where more appropriate change of basis $\Psi$ enable signal recovery with fewer measurements, accelerating the data collection process. Pre-defined change of basis such as the Discrete Cosine Transform (DCT) are traditionally used to form the matrix $\Psi$, but these basis are meant to be generic and are not tailored to the specific signal characteristics that can be encountered in each specific application. Hence, in order to make use of more appropriate change of basis, Dictionary Learning has been proposed as a way to \textit{learn} highly-specific change of basis matrices from a dataset of samples $\Bar{s}_i, i = 1,...,n_d$ representing the type of data that needs to be recovered by CS.  
\subsubsection{Dictionary Learning} 
\label{dicolearm}
(DL) is concerned with the \textit{unsupervised} learning of an appropriate change of basis matrix $\Psi$ (or dictionary) that best represents (in terms of mean square error minimization) the feature characteristics of signals of interest $\Bar{s}_i, i = 1,...,n_d$, subject to a sparsity constraint on the associated transform coefficients $\Bar{c}_i, i = 1,...,n_d$. DL learns $\Psi$ by solving the following joint optimization problem:
\begin{equation}
    \Bar{c}_i^*, \Psi^* = \arg \min_{\Bar{c}, \Psi} \sum_{i=1}^{n_d} \frac{1}{2} || \Psi \Bar{c}_i - \Bar{s}_i||_2^2 + \lambda ||\Bar{c}_i||_1
    \label{dicolearn}
\end{equation}
where $\lambda$ is a hyperparameter setting the sparsity level of the transform coefficients $\Bar{c}_i, \forall i$.

The joint optimization problem (\ref{dicolearn}) is typically solved by alternating between: \textit{i)} an optimization step that considers $\Psi$ fixed in (\ref{dicolearn}) and solves for all $\Bar{c}_i$, and \textit{ii)} an optimization step that considers all $\Bar{c}_i$ fixed in (\ref{dicolearn}) and solves for $\Psi$. After convergence, a dictionary or change of basis $\Psi^*$ is obtained that represents in a more faithful way the characteristics of the signal source compared to the use of pre-defined basis such as DCT and polynomial dictionaries \cite{polycs}.

In the next Section, DL and CS will be leveraged to set up a novel path planning method for guiding unmanned robot vehicles that can be used for environmental sensing tasks.

\section{Methods}
\label{method}

In this section, we present our novel Monte-Carlo-based approach for efficient path planning of autonomous sensing robots using Compressed Sensing (CS) and Dictionary Learning (DL). The key insight of our method is to leverage the structure of the measurement matrix $\Phi$ in the CS framework to generate optimized robot path plans that balance three critical factors: minimizing the robot's travel distance, maximizing the incoherence between the measurement matrix $\Phi$ and the dictionary $\Psi$, and ensuring high-quality signal reconstruction (i.e., a low reconstruction error).

\subsection{Problem Formulation}

Given an environmental area that needs to be sensed with high spatial resolution (e.g., a geographical region for $NO_2$ pollutant concentration measurement), our objective is to design an efficient path for a robot to follow while collecting measurements. The path should enable reconstruction of the entire field with high accuracy using CS while minimizing the robot's travel distance. We formulate this as an optimization problem within the CS framework, where $\Phi$ represents the \textit{binary} random measurement matrix defining which locations are measured (1's) and which are skipped (0's) and $\Psi$ represents the dictionary (or change of basis) learned from the training data. The robot path is then determined by finding the shortest path that traverses all the measurement locations (i.e., entries in $\Phi$ which contain values of $1$).

\subsection{Proposed Monte-Carlo Optimization}

Algorithm \ref{algo} details our proposed Monte-Carlo sampling approach which is used to optimize the design of the measurement matrix $\Phi$ within our CS setting. The proposed method generates random measurement matrix candidates and evaluates them based on a multi-objective cost function $C$ in (\ref{costfun}) that considers three contributions: \textit{i)} the path length $\mathcal{L}$ as the total distance the robot must travel to visit all measurement locations (the lower the better); \textit{ii)} the mutual incoherence $\mu$ (\ref{mutualincoh}) between the measurement matrix $\Phi$ and the dictionary $\Psi$, capturing the \textit{maximum inner product} between columns of $\Phi$ and $\Psi$, and which should be minimized following CS theory \cite{mutualincoh}; and \textit{iii)} the reconstruction error $\mathcal{E}$ computed on the validation dataset (the lower the better).
\begin{equation}
    C=\lambda_{valid} \cdot \mathcal{E} + \lambda_{path} \cdot \mathcal{L} + \lambda_{incoh} \cdot \mu
    \label{costfun}
\end{equation}

\begin{equation}
    \mu(\Phi, \Psi) = \max_{1 \leq q, r \leq n} \left| \langle \phi_q, \psi_r \rangle \right|
    \label{mutualincoh}
\end{equation}

Each contribution to the total cost $C$ in (\ref{costfun}) is weighted through hyper-parameters $\lambda_{valid}, \lambda_{path}$ and $\lambda_{incoh}$ which respectively set the importance of the reconstruction error $\mathcal{E}$, the path length $\mathcal{L}$ and the incoherence $\mu$.

For each generated $r^{th}$ random sample $\Phi^r$, Algorithm \ref{algo} evaluates its associated cost $C^{r}$ and if it is smaller than the previous minimum cost $C_{min}$, the newly generated sample $\Phi^r$ is kept as the current most optimal one $\Phi^* \xleftarrow{} \Phi^r$. After a number of $n_{iter}$ iterations, the measurement matrix $\Phi^*$ that has exhibited the smallest cost $C$ during the Monte-Carlo sampling is returned as the result of the optimization process.

%For each candidate matrix, we compute these metrics and select the one that minimizes the combined cost.

%Algorithm \ref{algo} presents the pseudocode for our approach:

\begin{algorithm}
 \caption{Monte-Carlo CS-DL Path Planning for Sensing Robots}
 \label{algo}
 \begin{algorithmic}[1]
 \renewcommand{\algorithmicrequire}{\textbf{Input:}}
 \renewcommand{\algorithmicensure}{\textbf{Output:}}
 \REQUIRE Dictionary $\Psi$ (e.g., learned through Dictionary Learning, see Eq. \ref{dicolearn}), validation data $\mathcal{X}_{valid}$, measurement matrix dimension $M \times N$, measurement sparsity $p$, number of Monte-Carlo samples $n_{iter}$, hyper-parameters $\lambda_{valid}$, $\lambda_{path}$, $\lambda_{incoh}$
 \ENSURE Optimized measurement matrix $\Phi^*$ and corresponding robot path
 \STATE Initialize $C_{min} \leftarrow \infty$ (initial minimum cost)
 \FOR{$r = 1$ to $n_{iter}$}
    \STATE Generate random binary matrix $\Phi^r \in \{0,1\}^{M \times N}$ with probability $p$ of having a $1$  and $(1-p)$ for $0$.
    \STATE // $\Phi^r$ indicates the $r^{th}$ sample
    \STATE Optionally: prune columns with few measurements ($\sum_j \Phi_{ij} \leq \theta$ with e.g., $\theta=3$)
    
    \STATE // Compute incoherence following (\ref{mutualincoh}).
    \STATE $\mu \leftarrow$ compute\_incoherence($\Phi^r$, $\Psi$)
    
    \STATE // Compute robot path length
    \STATE Convert $\Phi^r$ to binary measurement image $I$ by summing rows, thresholding (\ref{thresho}) and un-flattening (see Section \ref{robopathsec}).
    \STATE path$^r$, length$^r \leftarrow$ find\_shortest\_path\_over\_ones($I$)
    
    \STATE // Compute validation error
    \STATE error$^r \leftarrow 0$ // Initialize
    \FOR{each sample $\Bar{s}_j$ in $\mathcal{X}_{valid}$}
        \STATE $\Bar{s}_{m,j} \leftarrow \Phi^r \Bar{s}_j$ (obtain measurements)
        \STATE Solve $\min_{\Bar{c}} \frac{1}{2}\|\Phi^r \Psi \Bar{c} - \Bar{s}_{m,j}\|_2^2 + \lambda\|\Bar{c}\|_1$ using LASSO
        \STATE $\hat{s}_j \leftarrow \Psi \Bar{c}$ (reconstruct signal)
        \STATE error$^r \leftarrow$ error$^r + \|\Bar{s}_j - \hat{s}_j\|_2^2$
    \ENDFOR
    \STATE error$^r \leftarrow$ error$^r / |\mathcal{X}_{valid}|$ (average error)
    
    \STATE // Compute total cost
    \STATE $\mathcal{E} \xleftarrow{} $ error$^r$, $\mathcal{L} \xleftarrow{} $ length$^r$
    \STATE $C^r = \lambda_{valid} \cdot$ $\mathcal{E} + \lambda_{path} \cdot$ $\mathcal{L} + \lambda_{incoh} \cdot \mu$ // See (\ref{costfun}).
    
    \IF{$C^r < C_{min}$}
        \STATE $C_{min} \leftarrow C^r$
        \STATE $\Phi^* \leftarrow \Phi^r$
        \STATE $\text{path}^* \leftarrow \text{path}^r$
    \ENDIF
 \ENDFOR
 \RETURN $\Phi^*$, path$^*$
 \end{algorithmic} 
\end{algorithm}

\subsection{Robot Path Computation}
\label{robopathsec}
In order to derive the robot path from the measurement matrix $\Phi^*$, we remark that the measured signal $\Bar{s}_{m}$ is obtained from the original signal to be recovered $\Bar{s}$ as:
\begin{equation}
    \Bar{s}_{m} = \Phi^* \Bar{s}
    \label{meas}
\end{equation}

By nothing $N$ the dimension of $\Bar{s}$, we see in (\ref{meas}) that the entries of $\Bar{s}$ that need to be measured will correspond to the union of all the row-wise elements of $\Phi^*$ that are equal to $1$:
\begin{equation}
    \mathbf{\Bar{1}} = \{\Phi^*_1 \stackrel{?}{=} 1 \} \cup \{\Phi^*_2 \stackrel{?}{=} 2 \} \cup \hdots \cup \{\Phi^*_D \stackrel{?}{=} 1 \}
    \label{unions}
\end{equation}
where $\Phi^*_j$ denotes the $j^{th}$ row and $\mathbf{\Bar{1}}$ is an indicator vector containing 1's in the entries of $\Bar{s}$ that will be measured and 0's elsewhere.

In practice, (\ref{unions}) can be computed by first summing $\Phi^*$ along the columns and then thresholding the entries (line 8 in Algorithm \ref{algo}):
\begin{equation}
    \mathbf{\Bar{1}} = \{\Bar{a} \stackrel{?}{>} 0\} \hspace{3pt} \text{with} \hspace{3pt}  \Bar{a} = \sum_{j=1}^{M} \Phi^*_j
    \label{thresho}
\end{equation}

Now, assuming that the signal to be measured $\Bar{s}$ corresponds to a 2-dimensional heatmap that has been flatten as a 1-dimensional vector, the 2-dimensional coordinates that need to be measured on the heatmap grid can be obtained by first \textit{un-flattening} $\mathbf{\Bar{1}}$ back to a 2-dimensional representation and selecting the entries corresponding to values of $1$. 

Finally, the shortest path enabling the robot to cover all grid coordinates that need to be measured can be found using e.g., a greedy nearest-neighbor approach that approximate the solution to the Traveling Salesman Problem (TSP) \cite{nearestneigh} (the specific choice of path planning minimization algorithm does not have a significant importance in our context since it is applied offline during the optimization process of $\Phi$).

%A critical component of our approach is computing the robot's path length for a given measurement matrix. We transform the measurement matrix into a binary image where each pixel represents a location in the environment. The robot must visit all locations marked with a '1' (measurements to be taken) while minimizing travel distance.

%To solve this path planning problem, we use a greedy nearest-neighbor approach to approximate the solution to the Traveling Salesman Problem (TSP):

%\begin{enumerate}
%    \item Start from an arbitrary measurement location
%    \item Repeatedly move to the nearest unvisited measurement location
%    \item Continue until all measurement locations have been visited
%\end{enumerate}

%This provides a reasonably efficient path that the robot can follow to collect all required measurements.

\subsection{Enhanced Sparsification Method}
\label{enhancedspars}
As the sparsity of $\Phi$ has a crucial impact on the robot path length and measurement time, we propose a novel method within our CS context to enhance the sparsity level of the measurement matrix $\Phi$. First, our method counts the number of 1's along each column of $\Phi$. Then, for each column $i$ that has a number $n_{ones}$ of 1's smaller than a threshold $\theta$, all entries in this column are forced to zero, effectively reducing the number of 1's further:
\begin{equation}
    \Phi^*_{ij} = 0, \forall j = 1,...,D  \hspace{3pt} \text{if} \hspace{3pt} a_i < \theta \hspace{3pt} \text{with} \hspace{3pt} a_i = \sum_{j=1}^{M} \Phi^*_{ij}
    \label{sparshenc}
\end{equation}

In Algorithm \ref{algo}, the sparsity enhancement step (\ref{sparshenc}) is optionally applied in line 4. It will be shown during our experiments that this proposed method for sparsity enhancement enables a significant reduction in robot path length while not significantly affecting the signal reconstruction quality.

\section{Experimental Results}
\label{results}

\subsection{Setup}
\begin{figure}[htbp]
    \centering
    \includegraphics[width=0.4\textwidth]{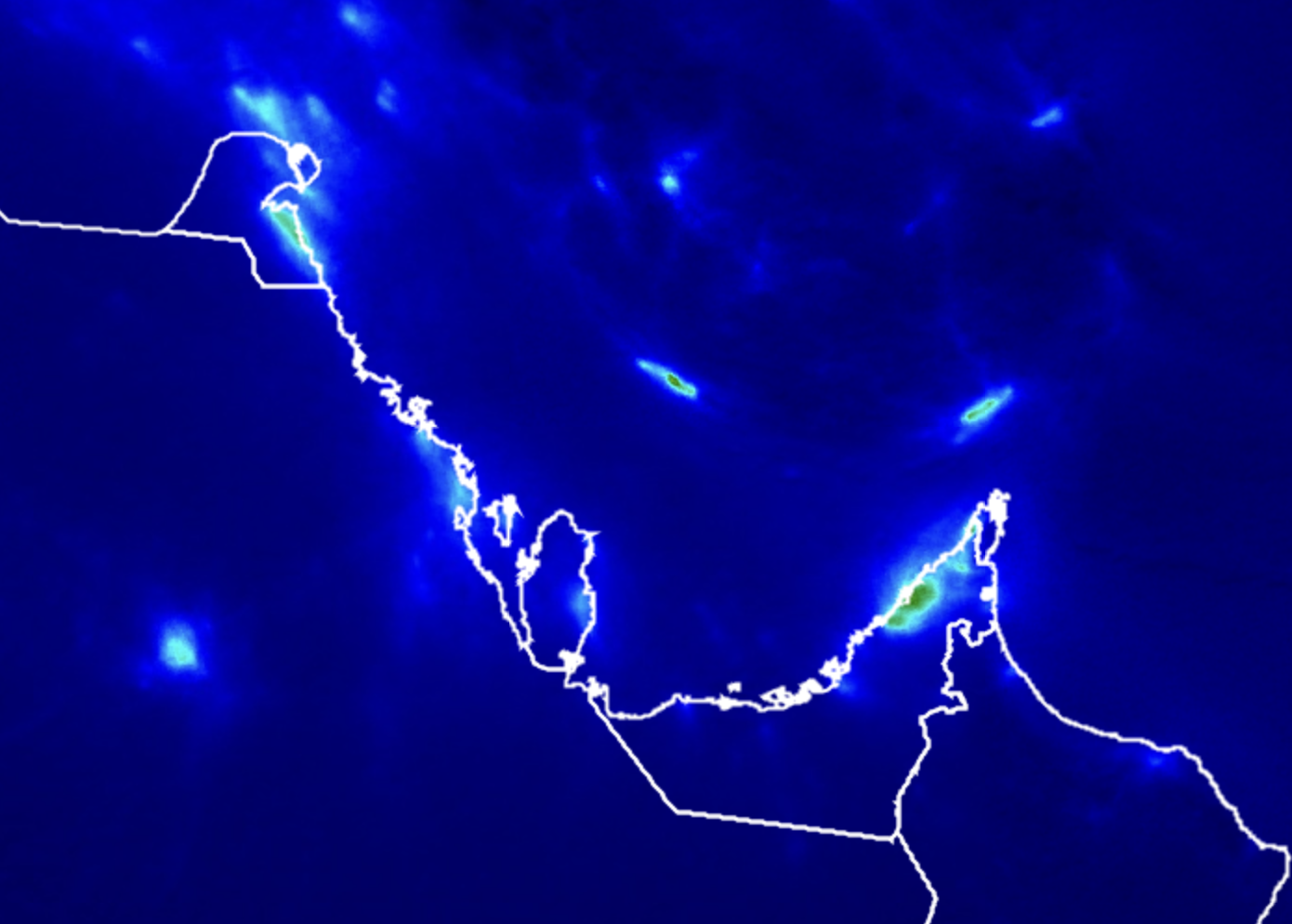} % Adjust the width percentage
    \caption{\textit{\textbf{Nitrogen dioxide ($NO_2$) concentration heatmap} used as data source during our experiments. Brighter values indicate a higher $NO_2$ concentration. Values on the map are normalized between $0$ and $1$. Countries' borders are shown for illustration purposes.}}
    \label{datasource}
\end{figure}

In our experimental setup, we utilize environmental maps that illustrate atmospheric concentrations of $NO_2$ pollutants across the Gulf region adjacent to the state of Qatar, where our research group operates. The $NO_2$ data is obtained via Google Earth and is based on satellite imagery from the Copernicus Sentinel-2 mission. Fig. \ref{datasource} displays the selected geographical region and its associated $NO_2$ concentration heatmap ($1390 \times 994$ pixels), which serves as the input dataset in our study. %The map resolution used in the experiments is $1390 \times 994$ pixels.

Using the heatmap shown in Fig. \ref{datasource} as the input dataset, we apply a sliding window approach with a stride of $32$ to extract image patches of size $128 \times 128$. This process yields a total of $1120$ $NO_2$ concentration patches that are spatially distributed across the defined geographical region. To reduce the computational cost associated with our experiments, the extracted $128 \times 128$ patches are subsequently down-sampled to dimensions of $32 \times 32$. As a result, we construct a dataset denoted by $\mathcal{D} = { P_i, i = 1, ..., n_d }$, where each $P_i$ represents a $32 \times 32$ patch of $NO_2$ concentration values, and the total number of samples is $n_d = 1120$.

In the following experiments, we partition the dataset $\mathcal{D}$ into three subsets: $70\%$ for training, $5\%$ for validation, and $25\%$ for testing. The training subset is employed to learn a dictionary $\Psi$, which serves as the sparsifying basis, using the DL approach detailed in Section \ref{dicolearm}. This step is implemented using the \texttt{sklearn} library in \textit{python}. After learning $\Psi$, we utilize the validation subset to perform Monte Carlo-based optimization (outlined in Algorithm \ref{algo}) in order to derive an optimized measurement matrix $\Phi^*$ along with the corresponding robot path length for data collection. Finally, the learned dictionary $\Psi$ and the optimized matrix $\Phi^*$ are used in the DL-CS framework to evaluate the map reconstruction performance on the test subset.

Subsequently, we employ the DL-CS-based path planning approach described in Section \ref{method} on the dataset $\mathcal{D}$ constructed in this section and present the corresponding experimental outcomes.

\subsection{Results}

\begin{figure}[htbp]
    \centering
    \includegraphics[width=0.52\textwidth]{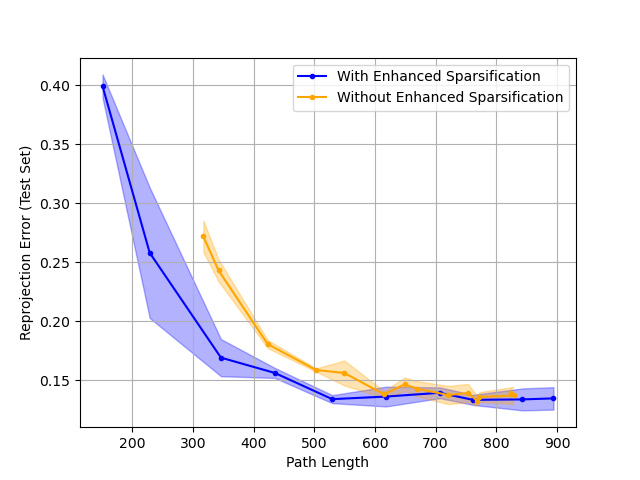} % Adjust the width percentage
    \caption{\textit{\textbf{CS reconstruction error on the test set in function of the robot path length.} It can be clearly remarked that using the proposed Enhanced Sparsification method (\ref{sparshenc}), a significantly lower reconstruction error is obtained for a given path length. Both curves converge to a similar reconstruction error $\sim0.14$ as the path length grows. In addition, using the proposed Enhanced Sparsification method (\ref{sparshenc}) enables the Monte-Carlo procedure to reach lower robot path lengths $\sim200$ during its optimization process vs. without using (\ref{sparshenc}) in the orange graph.}}
    \label{res11}
\end{figure}

Fig. \ref{res11} illustrates the CS reconstruction error on the test set as a function of the robot’s path length. These results are obtained by executing Algorithm \ref{algo} for varying values of the probability parameter $p = {0.03, 0.04, ..., 0.12}$, which controls the likelihood of generating a 1 during the random construction of the measurement matrix $\Phi$ (see line 3 in Algorithm \ref{algo}). For comparison, the figure also includes the outcomes obtained without incorporating the proposed Enhanced Sparsification (ES) technique introduced in Section \ref{enhancedspars}. Each plot is averaged over three independent trials, each using a distinct random initialization (kept consistent across all experiments). The reported curves represent the mean performance across these runs, with the corresponding standard deviation indicated by the shaded regions.

As anticipated, Fig. \ref{res11} demonstrates that an increase in the robot’s path length leads to a reduction in reconstruction error on the test set, due to the larger number of signal measurements collected along longer paths. Additionally, it can be observed that both curves reach a plateau at path lengths of approximately $\sim 600$, where the reconstruction error stabilizes around $\sim 0.14$.

Notably, Fig. \ref{res11} highlights the substantial benefit of incorporating our ES technique introduced in Section \ref{enhancedspars}, which results in approximately a $60\%$ reduction in CS reconstruction error for a given robot path length. For example, at a path length of around $\sim 300$, applying ES yields a test error of roughly $\sim 0.17$, compared to $\sim 0.275$ when the method is not applied (as shown by the orange curve). These findings clearly underscore the effectiveness of the proposed ES approach in generating optimized CS measurement matrices $\Phi$, enabling shorter robot paths while achieving high-precision data reconstruction (i.e., lower test errors).

\begin{figure}[htbp]
    \centering
    \includegraphics[width=0.5\textwidth]{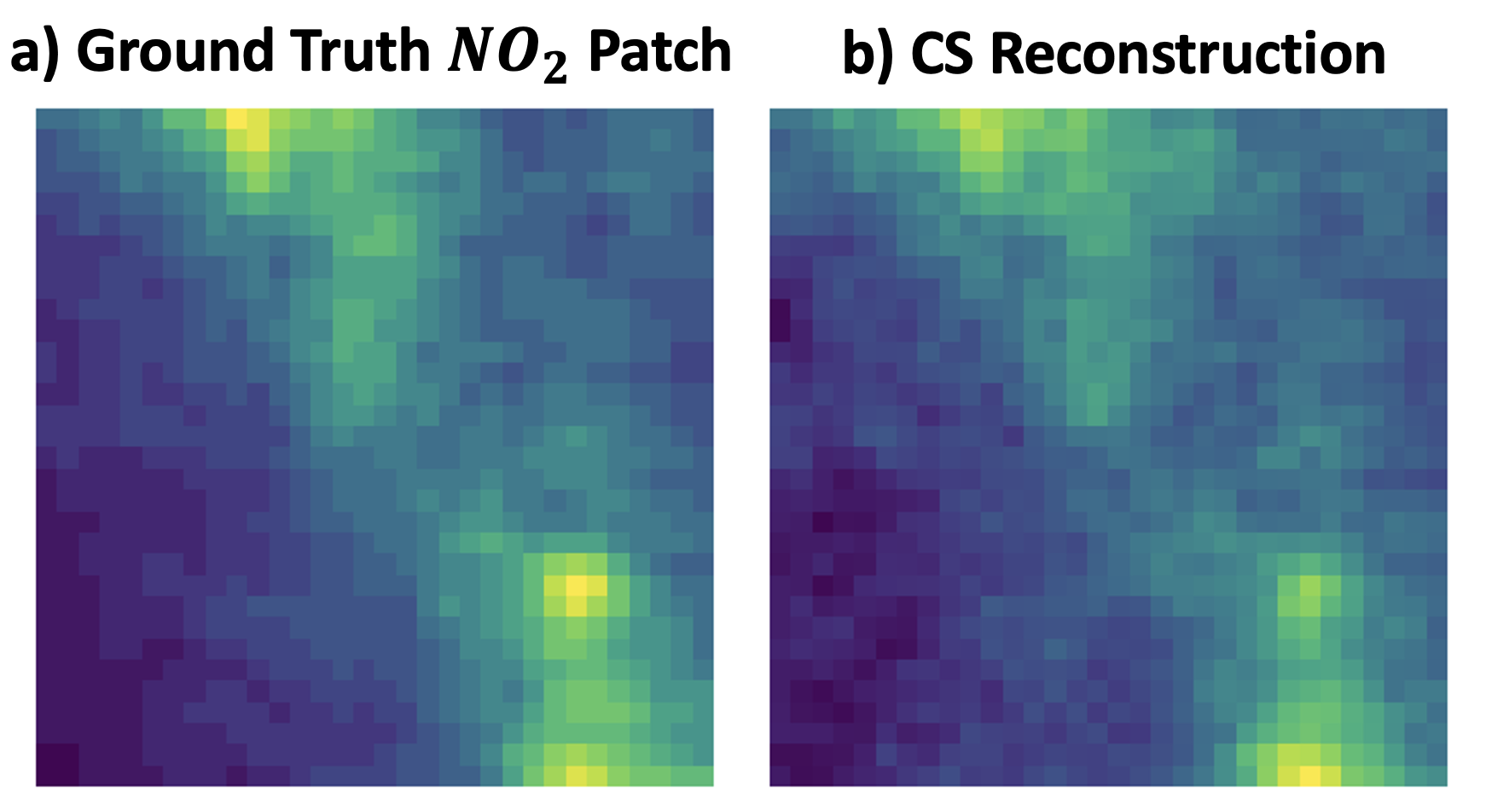} % Adjust the width percentage
    \caption{\textit{\textbf{Example of $NO_2$ map reconstruction using our proposed Monte-Carlo CS measurement matrix optimization.} a) Ground-truth $NO_2$ concentration map ($32\times32$) in which the robot is set to navigate and collect compressed measurements; b) Reconstructed map obtained using the optimized robot measurement path. }}
    \label{compar}
\end{figure}

\begin{figure}[htbp]
    \centering
    \includegraphics[width=0.36\textwidth]{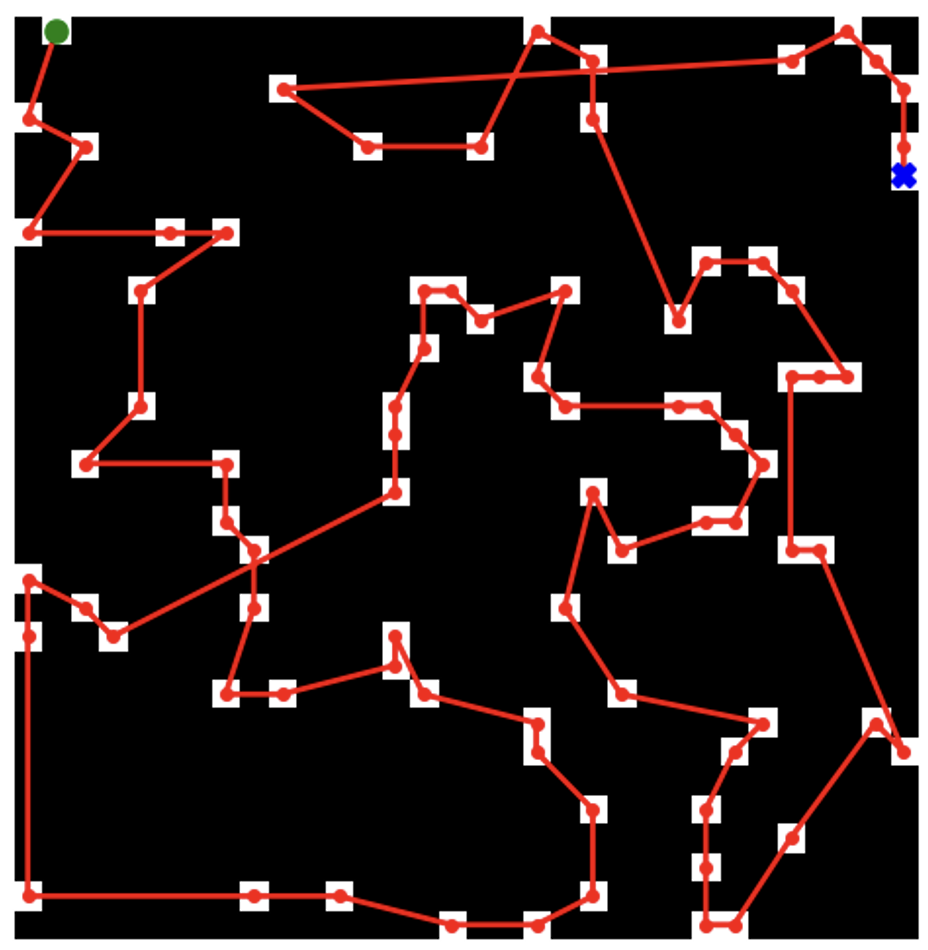} % Adjust the width percentage
    \caption{\textit{\textbf{Optimized robot sensing path.} The area to be sensed is subdivided into a $32\times32$ grid.}}
    \label{pathrobot}
\end{figure}

To illustrate the effectiveness of our approach, Fig. \ref{compar} presents a reconstructed $NO_2$ concentration map obtained using the optimized measurement matrix $\Phi^*$ along with the corresponding robot data collection trajectory, shown in Fig. \ref{pathrobot}, as produced by Algorithm \ref{algo}. As depicted in Fig. \ref{pathrobot}, the resulting robot path is notably sparse, requiring visits to less than $10\%$ of the total locations within the $32 \times 32$ grid, while still enabling an accurate reconstruction of the concentration map.

\subsection{Effect of the hyper-parameters on the optimization results}

\begin{figure}[htbp]
    \centering
    \includegraphics[width=0.51\textwidth]{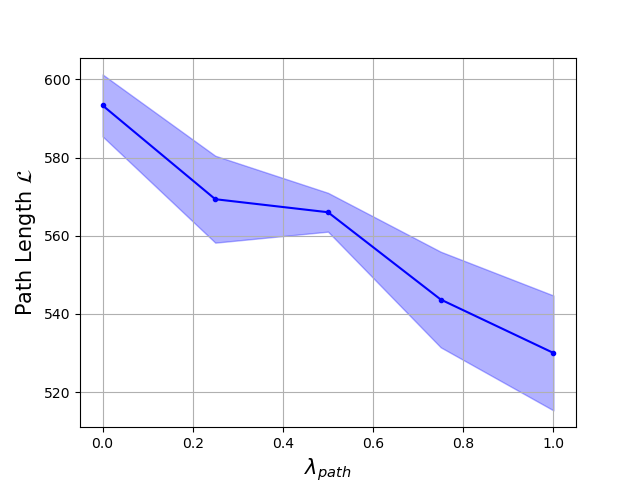} % Adjust the width percentage
    \caption{\textit{\textbf{Effect of $\lambda_{path}$ on the generated robot path length.} }}
    \label{path1}
\end{figure}

\begin{figure}[htbp]
    \centering
    \includegraphics[width=0.51\textwidth]{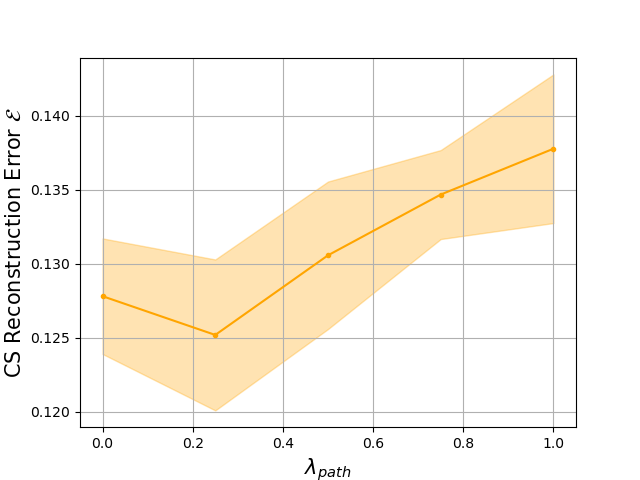} % Adjust the width percentage
    \caption{\textit{\textbf{Effect of $\lambda_{path}$ on CS reconstruction error.} }}
    \label{err12}
\end{figure}

Additionally, we investigate how the optimization hyper-parameters in (\ref{costfun}) influence the resulting robot path length, reconstruction error, and CS incoherence metrics. In the following experiments, we fix $\lambda_{valid} = 1$ and systematically vary the values of $\lambda_{path}$ and $\lambda_{incoh}$ to analyze the impact of different hyper-parameter configurations. Fig. \ref{path1} illustrates how changes in $\lambda_{path}$ affect the generated robot path length. As expected, increasing $\lambda_{path}$ results in shorter robot trajectories. Fig. \ref{err12} further shows the corresponding effect on CS reconstruction error: as $\lambda_{path}$ increases, the test error also rises. This behavior is anticipated, as higher values of $\lambda_{path}$ emphasize minimizing path length, thereby reducing the number of collected measurements and ultimately degrading reconstruction accuracy.

\begin{figure}[htbp]
    \centering
    \includegraphics[width=0.51\textwidth]{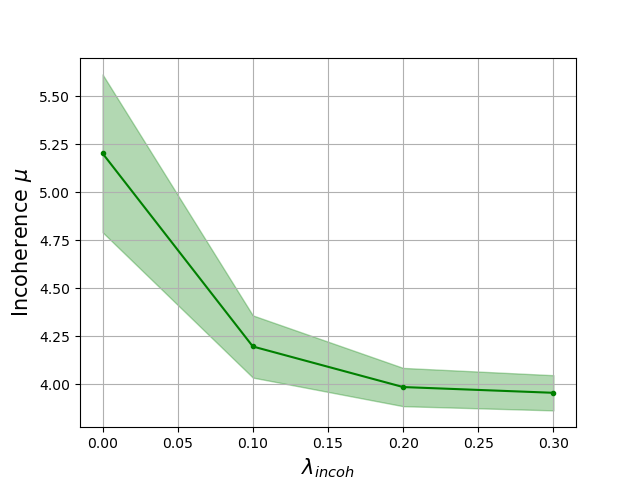} % Adjust the width percentage
    \caption{\textit{\textbf{Effect of $\lambda_{incoh}$ on the incoherence $\mu$ of the optimized measurement matrix $\Phi^*$.} }}
    \label{incoh3}
\end{figure}

Regarding the influence of $\lambda_{incoh}$, Fig. \ref{incoh3} demonstrates that increasing $\lambda_{incoh}$ results in a reduction of the incoherence measure $\mu$ (defined in Equation \ref{mutualincoh}) between the optimized measurement matrix $\Phi^*$ and the dictionary $\Psi$. Although no statistically significant impact of $\lambda_{incoh}$ on the reconstruction test error was observed using our dataset, adjusting this hyper-parameter allows for explicit control over the incoherence $\mu(\Phi^*, \Psi)$. This capability may prove valuable in certain application contexts, given that minimizing incoherence is well-established in CS theory as a key factor for successful signal recovery \cite{cspaper2,mutualincoh}.

\subsection{Dictionary Learning vs. pre-defined dictionaries}
\begin{figure}[htbp]
    \centering
    \includegraphics[width=0.52\textwidth]{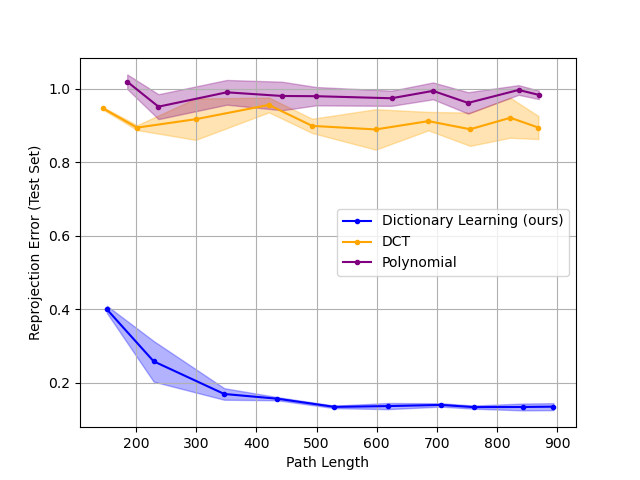} % Adjust the width percentage
    \caption{\textit{\textbf{Comparing the use of DL with pre-defined DCT and Polynomial dictionaries.} Using DL to learn a well-suited change of basis $\Psi$ (as used in our proposed method) leads to more than $\times 5$ lower CS reconstruction error.}}
    \label{vsdct}
\end{figure}

In addition, Fig. \ref{vsdct} explores the significance of employing DL, as described in Section \ref{dicolearm}, to obtain an adapted basis $\Psi$, in contrast to relying on standard, pre-defined dictionaries like the Discrete Cosine Transform (DCT) or polynomial dictionaries, which are commonly used in compressed sensing literature \cite{polycs}.

Fig. \ref{vsdct} clearly demonstrates that employing dictionary learning (DL), as implemented in our proposed framework, is essential for achieving low reconstruction errors. While pre-defined dictionaries such as the DCT and polynomial bases are widely used in compressed sensing, they fail to deliver satisfactory reconstruction accuracy when applied with low-dimensional measurement matrices that correspond to short robot path lengths, as considered in this study. This finding aligns with prior work in the literature \cite{sparselearning}, which emphasizes the advantages of \textit{learned} dictionaries in enabling high-precision compressed sensing reconstruction even when working with highly sparse measurement matrices containing few measurement dimensions (e.g., $M=35$ in our experiments).

%Interestingly, Fig. \ref{vsdct} shows that the test error for the Wavelet case grows as the path length grows. This effect seems to be caused by our use of measurement matrices with only $M=35$ measurement dimensions. Indeed, increasing $M$ towards values of $M\sim 1000$ led to significantly lower errors around $\sim0.35$ but consequently, leads to \textit{excessively long path lengths} around $\sim1000$. This confirms the crucial role of DL for achieving a low reconstruction error while using short robot measurement path lengths.

\subsection{Comparison to prior work}
\begin{figure}[htbp]
    \centering
    \includegraphics[width=0.52\textwidth]{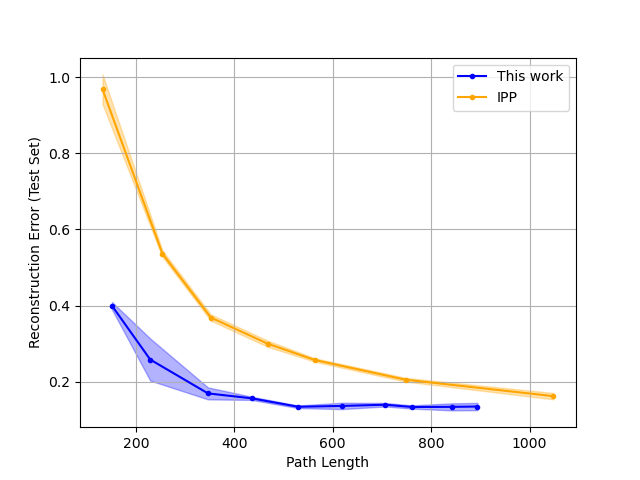} % Adjust the width percentage
    \caption{\textit{\textbf{Comparing our proposed method to IPP.} It can be clearly seen that our proposed approach significantly outperforms IPP in terms of reaching a lower reconstruction error for a given robot path length (by more than $\times 2$ lower error on average).}}
    \label{comp_IPP}
\end{figure}

Finally, it is important to compare our proposed method against the closely-related IPP method \cite{infopathplan, infogad} for robot path planning and map recovery from a reduced set of measurements. Even though the goal of IPP and our proposed method is similar, our approach fundamentally differs from IPP through the use of CS and DL, while IPP makes use of Gaussian Processes. In order to compare our approach against IPP, we apply IPP to the 3-fold test set data used in the experiments throughout this Section, and we compute the average reconstruction error between the ground-truth samples and the recovered maps. We repeat these experiments with a varying number of sensing points, leading to different robot path lengths. Fig. \ref{comp_IPP} compares the performance of our proposed approach with IPP.

Fig. \ref{comp_IPP} clearly shows that our proposed approach significantly outperforms IPP by reaching more than $\times 2$ lower reconstruction error on average for a given robot path length. The ability of our approach to reach a lower reconstruction error is expected to be due to the use of DL, which learns a well-suited dictionary $\Psi$ while imposing a sparsity constraint on the signal representations during learning (through the $l_1$ norm in Eq. \ref{dicolearn}). In contrast, IPP uses Gaussian Process modeling which does not impose a sparse prior on the data representation, necessitating more measurements to reach a low reconstruction error.

Finally, another key aspect in which our proposed approach outperforms IPP is related to the runtime \textit{compute complexity} of both approaches. IPP performs path planning in real time, where robots need to maintain a Gaussian Process model of the environment in which they navigate. For each path planning step, the robot needs to find a direction of navigation that maximizes the information gain of the Gaussian Process. This is done in practice by executing an \textit{online} gradient descent procedure each time the robot is moving towards its next sensing location. In contrast, our proposed algorithm does not require any compute-expensive online optimization step. Our approach effectively pushes the compute-expensive aspects such as DL and the Monte-Carlo optimization of $\Phi$ (Algorithm \ref{algo}) into a fully \textit{offline} phase. Once this offline phase is executed (e.g., on a back-end server), the optimized sensing matrix $\Phi^*$ is directly used to direct the robot path, dropping the requirement for compute-expensive \textit{online} optimization at the robot side. This advantage is particularly important in the case of resource-constrained robot platforms such as drones which feature limited battery life and limited compute power.

\section{Conclusion}
\label{conclus}
This paper has introduced a novel Monte Carlo-based method for optimizing measurement matrix design in Compressed Sensing applications, aiming to achieve high-accuracy signal reconstruction while minimizing the robot’s navigation path length required to gather the compressed measurements. Importantly, our results demonstrate that leveraging dictionary learning to obtain the sparsifying basis significantly reduces $NO_2$ map reconstruction errors for a given robot path length, outperforming conventional approaches that rely on pre-defined dictionaries such as DCT and polynomial bases, which tend to produce higher reconstruction errors. In addition, our approach significantly outperforms the previously-proposed Information Path Planning (IPP) method by reaching more than $\times 2$ lower reconstruction errors for a given path length. As future work, we intend to pursue the deployment of our path planning strategy on actual robotic data collection platforms and showcase its application in practical scenarios.

% or


%\section*{Acknowledgment}


\ifCLASSOPTIONcaptionsoff
  \newpage
\fi


\begin{thebibliography}{1}

\bibitem{ocean1} A. Astolfi et al., "Marine Sediment Sampling With an Underwater Legged Robot: A User-Driven Sampling Approach for Microplastic Analysis," in IEEE Robotics \& Automation Magazine, vol. 31, no. 1, pp. 62-71, March 2024, doi: 10.1109/MRA.2023.3341288. 

\bibitem{ocean2} N. A. Cruz and A. C. Matos, "The MARES AUV, a Modular Autonomous Robot for Environment Sampling," OCEANS 2008, Quebec City, QC, Canada, 2008, pp. 1-6, doi: 10.1109/OCEANS.2008.5152096.

\bibitem{air1} N. Karna, M. A. M. Firdausa and S. Y. Shin, "Air Quality Index Mapping Using Programmable Single Propeller UAV Towards Internet of Drone Things," 2023 14th International Conference on Information and Communication Technology Convergence (ICTC), Jeju Island, Korea, Republic of, 2023, pp. 805-810, doi: 10.1109/ICTC58733.2023.10393444.

\bibitem{air2} G. M. Bolla et al., "ARIA: Air Pollutants Monitoring Using UAVs," 2018 5th IEEE International Workshop on Metrology for AeroSpace (MetroAeroSpace), Rome, Italy, 2018, pp. 225-229, doi: 10.1109/MetroAeroSpace.2018.8453584.

\bibitem{rover1} M. R. Ershadi et al., "Autonomous Rover Enables Radar Profiling of Ice-Fabric Properties in Antarctica," in IEEE Transactions on Geoscience and Remote Sensing, vol. 62, pp. 1-9, 2024, Art no. 5913809, doi: 10.1109/TGRS.2024.3394594.

\bibitem{rover2} A. Holloway et al., "The Myth of the Data-Constrained Mission: Ten Years of Data Management Onboard the Curiosity Rover," 2024 IEEE Aerospace Conference, Big Sky, MT, USA, 2024, pp. 1-16, doi: 10.1109/AERO58975.2024.10521196.

\bibitem{infopathplan} R. Marchant and F. Ramos, "Bayesian Optimisation for informative continuous path planning," 2014 IEEE International Conference on Robotics and Automation (ICRA), Hong Kong, China, 2014, pp. 6136-6143, doi: 10.1109/ICRA.2014.6907763.

\bibitem{infogad} Hollinger GA, Sukhatme GS. Sampling-based robotic information gathering algorithms. The International Journal of Robotics Research. 2014;33(9):1271-1287. doi:10.1177/0278364914533443

\bibitem{dronex1} A. Safa et al., "FMCW Radar Sensing for Indoor Drones Using Variational Auto-Encoders," 2023 IEEE Radar Conference (RadarConf23), San Antonio, TX, USA, 2023, pp. 1-6, doi: 10.1109/RadarConf2351548.2023.10149738.

\bibitem{dronex3} A. Safa et al., "Learning to Encode Vision on the Fly in Unknown Environments: A Continual Learning SLAM Approach for Drones," 2022 IEEE International Symposium on Safety, Security, and Rescue Robotics (SSRR), Sevilla, Spain, 2022, pp. 373-378, doi: 10.1109/SSRR56537.2022.10018713.

\bibitem{daniel} M. D. Alea et al., "A Fingertip-Mimicking 12 × 16 200  $\mu$m-Resolution e-Skin Taxel Readout Chip With Per-Taxel Spiking Readout and Embedded Receptive Field Processing," in IEEE Transactions on Biomedical Circuits and Systems, vol. 18, no. 6, pp. 1308-1320, Dec. 2024, doi: 10.1109/TBCAS.2024.3387545.

\bibitem{cspaper} D. L. Donoho, "Compressed sensing," in IEEE Transactions on Information Theory, vol. 52, no. 4, pp. 1289-1306, April 2006, doi: 10.1109/TIT.2006.871582.

\bibitem{cspaper2} E. J. Candes and M. B. Wakin, "An Introduction To Compressive Sampling," in IEEE Signal Processing Magazine, vol. 25, no. 2, pp. 21-30, March 2008, doi: 10.1109/MSP.2007.914731. 

\bibitem{dictionarylearning} I. Tošić and P. Frossard, "Dictionary Learning," in IEEE Signal Processing Magazine, vol. 28, no. 2, pp. 27-38, March 2011, doi: 10.1109/MSP.2010.939537.

\bibitem{nguyen2020} M. T. Nguyen and H. R. Boveiri, "Energy-efficient sensing in robotic networks," Measurement, vol. 158, 2020, Art. no. 107708, doi: 10.1016/j.measurement.2020.107708.

\bibitem{yan2012} S. Yan, C. Wu, W. Dai, M. Ghanem and Y. Guo, "Environmental monitoring via compressive sensing," in Proceedings of the Sixth International Workshop on Knowledge Discovery from Sensor Data, Beijing, China, 2012, pp. 61–68, doi: 10.1145/2350182.2350189.

\bibitem{chang2023} Q. Chang, Y. Liu, Z. Wang, Y. Zhang and T. Abdelzaher, "Deep compressed sensing based data imputation for urban environmental monitoring," ACM Transactions on Sensor Networks, Early Access, 2023, doi: 10.1145/3599236.

\bibitem{liu2024} S. Liu, X. Wang, J. Song and Z. Han, "Cooperative Robotics Visible Light Positioning: An Intelligent Compressed Sensing and GAN-Enabled Framework," IEEE Journal of Selected Topics in Signal Processing, vol. 18, no. 3, pp. 407–418, Apr. 2024, doi: 10.1109/JSTSP.2024.3368661.

\bibitem{mutualincoh} R. Obermeier and J. A. Martinez-Lorenzo, "Sensing Matrix Design via Mutual Coherence Minimization for Electromagnetic Compressive Imaging Applications," in IEEE Transactions on Computational Imaging, vol. 3, no. 2, pp. 217-229, June 2017, doi: 10.1109/TCI.2017.2671398.

\bibitem{nearestneigh} Kizilateş, G., Nuriyeva, F. (2013). On the Nearest Neighbor Algorithms for the Traveling Salesman Problem. In: Nagamalai, D., Kumar, A., Annamalai, A. (eds) Advances in Computational Science, Engineering and Information Technology. Advances in Intelligent Systems and Computing, vol 225. Springer, Heidelberg. https://doi.org/10.1007/978-3-319-00951-3\_11

\bibitem{sparselearning} Brunton, B., Brunton, S., Proctor, J., \& Kutz, J. (2016). "Sparse Sensor Placement Optimization for Classification." SIAM Journal on Applied Mathematics, 76(5), 2099-2122.

\bibitem{polycs} Chkifa, A., Dexter, N., Tran, H. \& Webster, C. (2016). Polynomial approximation via compressed sensing of high-dimensional functions on lower sets. Mathematics of Computation. 87. 10.1090/mcom/3272. 

\end{thebibliography}
\end{document}